\documentclass[twoside,11pt]{article}

\usepackage{jmlr2e}

\usepackage{amsmath}
\usepackage{amsfonts}
\usepackage{microtype}
\usepackage{graphicx}
\usepackage{subcaption}
\usepackage{booktabs} 
\usepackage{xcolor}

\usepackage[linesnumbered, ruled,vlined]{algorithm2e}
\usepackage{mathtools}		

\usepackage{wrapfig}

\ewrlheading{14}{2018}{October 2018, Lille, France}{Mohammed Amin Abdullah, Aldo Pacchiano, Moez Draief}

\ShortHeadings{Reinforcement Learning with Wasserstein Regularisation}{Mohammed Amin Abdullah, Aldo Pacchiano, Moez Draief}
\firstpageno{1}

\begin{document}

\title{Reinforcement Learning with Wasserstein Distance Regularisation, with Applications to Multipolicy Learning}

\author{Mohammed Amin Abdullah\thanks{Huawei Technologies Ltd. {\tt mohammed.abdullah@huawei.com}, joint first author.}, Aldo Pacchiano\thanks{UC Berkeley, {\tt pacchiano@berkeley.edu}, joint first author.}, Moez Draief\thanks{Huawei Technologies Ltd. {\tt moez.draief@huawei.com}}}
%

\maketitle

\begin{abstract}
We describe an application of Wasserstein distance to Reinforcement Learning. The Wasserstein distance in question is between the distribution of mappings of trajectories of a policy into some metric space, and some other fixed distribution (which may, for example, come from another policy). Different policies induce different distributions, so given an underlying metric, the Wasserstein distance quantifies how different policies are. This can be used to learn multiple polices which are different in terms of such Wasserstein distances by using a Wasserstein regulariser. Changing the sign of the regularisation parameter, one can learn a policy for which its trajectory mapping distribution is attracted to a given fixed distribution. 
\end{abstract}


\section{Introduction and Motivation}
\noindent
Reinforcement learning (RL) is a formalism for to modelling and solving sequential decision problems in which an agent interacts with its environment and receives a scalar reward \citep{suton1998reinforcement}. In recent times, deep reinforcement learning has been successfully used to solve numerous continuous and discrete control tasks, often in continuous state spaces \citep{mnih2015human,silver2016mastering}. 

In this paper we consider the classical reinforcement learning problem of an agent that interacts with its environment while trying to maximise its cumulative reward \citep{burnetas1997optimal, kumar2015stochastic}. We are interested in RL problems modeled as Markov Decision Processes (MDP) $\mathcal{M} = (\mathcal{S}, \mathcal{A}, \mathcal{T}, R, \gamma) $ where $\mathcal{S}$ and $\mathcal{A}$ denote the state and action spaces which could be discrete or continuous, $\gamma \in (0,1]$ is the discount factor, and $\mathcal{T}$ and $R$ are the transition and reward functions respectively. The transition function $\mathcal{T}: \mathcal{S} \times \mathcal{A} \rightarrow \Delta_\mathcal{S}$ specifies the dynamics of the MDP and in general is assumed to be unknown. Where given any - possibly infinite- set $\mathcal{X}$, we denote by $\Delta_{\mathcal{X}}$ the set of distributions over $\mathcal{X}$. The reward function $R: \mathcal{S} \times \mathcal{A} \rightarrow \mathbb{R}$ specifies the utility gained by the agent in a given transition. In this paper we consider both finite and infinite horizon MDPs. Due to practical considerations for our experimental results we consider finite horizon problems.


In an MDP, a \emph{trajectory} $\tau = (s_0, a_0, r_0, s_1, a_1, r_1, \cdots )$ with $s_i \in \mathcal{S}$, $a_i \in \mathcal{A}$ and $r_i \in \mathbb{R}$ for all $i$ is a sequence encoding the visited states, actions taken and rewards obtained during an episode of the agent's interaction with the environment. All trajectories satisfy $\tau \in \left( \mathcal{S} \times \mathcal{A} \times \mathbb{R} \right)^H$, where $H$ is the MDP's horizon. For ease of notation we set $H = \infty$ in what follows, although all our results hold for finite $H$ and also for cases when the MDPs trajectories can have variable lengths.

Any MDP $\mathcal{M}$ has a corresponding set of trajectories $\Gamma \subset \left( \mathcal{S} \times \mathcal{A} \times \mathbb{R} \right)^\infty$ that may be taken. In this paper we consider stochastic policies that map any state in $\mathcal{S}$ to a distribution over actions. Any policy $\pi: \mathcal{S} \rightarrow \Delta_{\mathcal{A}}$ over $\mathcal{M}$ will induce a probability measure over its space of trajectories $\Gamma$. Let $M$ be a metric space endowed with a metric $d:M \times M \rightarrow \mathbb{R}_+$. An embedding is a function $f:\Gamma \rightarrow M$  mapping trajectories to points in $M$. For any MDP and policy pair $(\mathcal{M}, \pi)$, an embedding $f : \Gamma \rightarrow \mathbb{R}$ induces a distribution $\pi_M$ over $M$.  Different policies will induce different measures. 

The main aim of this paper is to tackle the problem of defining a similarity measure between different policies acting over the same MDP. The central contribution of this work is to propose the use of metric space embeddings for the trajectory distributions induced by a policy and use them as the basis of novel algorithms for policy attraction and policy repulsion. The principal ingredient of our proposed algorithms is the use of a computationally tractable alternative to the Wasserstein distance between the distributions induced by trajectory embeddings. The metric $d: M \times M \rightarrow \mathbb{R}_+$ gives us a way to quantify how much individual trajectories differ, and this in turn can be used in the framework of optimal transport~\citep{villani2008optimal} to give a measure of how different the \emph{behaviour} of different policies are. 

\noindent
There are many reasons why we might care about how the behaviour of policies compare against each other. For example, in the control of a telecommunications network, designing for robustness is central, and promotion of diversity is a classical approach \citep{pioro2004routing}. A single policy may be quite
heavily reliant on certain sub-structures (links, nodes, frequencies, etc.), but due to reliability issues, it may be desirable to have other policies at our disposal which perform well but which behave differently to each other. This way, there are back-up options if the parts of the network go down. That is, in learning the different policies, we wish the trajectory distributions to have a \emph{repulsive} effect on each other. This becomes especially pertinent with the rise of Software Defined Networking (SDN) \citep{kreutz2015software}, which, in a nutshell, is a paradigm in which the ``intelligent'' components  of network control (broadly speaking, the algorithms for resource allocation) are moved away from the routers into a (logically) centralised software controller. The routers become dumb but very fast machines which take their direction from the centralised controller. This present opportunities for online \citep{paris2016online} and real-time \citep{allybokus2017real} control, but naturally places a smaller time horizon for mitigating robustness problems. 

\noindent
Another motivation is the subject of the first of our experiments. Here, we wish to model an agent trying to learn to maneuver through most efficient route between designated start and end points over a hilly terrain (modelled in our experiments as a grid world \citep{suton1998reinforcement}). The most efficient route(s) maybe affected by the specifics of the agent itself, merely due to the physics of the scenario (e.g., consider the difference between an off-road 4x4 vehicle and a smaller delivery pod such as those of Starship Technologies ({\tt www.starship.xyz})), so the best route for one type of agent may not be exactly the best route for another. However, they may be similar, and this motivates us to consider using a pre-existing good route to influence the learning of the agent; the trajectory distributions have an \emph{attractive} effect on each other. Within our algorithm, the difference between attractive and the aforementioned repulsive effects is a change of sign in the regularisation factor of the Wasserstein term in an objective function. 

\noindent
Informally stated, our contributions are as follows: \emph{We present reinforcement learning algorithms for finding a policy $\pi^*$ where the objective function is the standard return plus regulariser that approximates the Wasserstein distance between the distribution of a mapping of trajectories induced by $\pi^*$ and some fixed distribution.} Thus, the algorithm tries to find a good (in the standard sense) policy $\pi^*$ whose trajectories are different or similar to some other, fixed, distribution of trajectories. 

\noindent
It is clear that in the end, the aim of control is not the policy itself but the actual behaviour of the system, which in reinforcement learning is the distribution over trajectories. The Wasserstein distance, also known as the \emph{Earth-Mover Distance} (EMD)~\citep{villani2008optimal, santambrogio2015optimal}, exploits an underlying \emph{geometry} that the the trajectories exist in, which is something that say,  Kullback-Leibler (KL) or Total Variation (TV) distance don't. This is advantageous when a relevant geometry can be defined. It is particularly pertinent to RL because different trajectories means different behaviours, and we would like to quantify how different the behaviours of two policies are in terms of how different are the trajectories that they take. For example, if $\Gamma=\{\tau_1, \tau_2,\tau_3\}$, and each of three policies $\pi_i$ induces Dirac on $\tau_i$, then in KL and TV terms, any pair of policies are just as different to each other as any other pair. However, if $\tau_1$ and $\tau_2$ are very similar in terms of behaviour (as defined by $d(f(\tau_1), f(\tau_2)$), but are both very different to $\tau_3$, then this will not be captured by KL and TV, but will be captured by Wasserstein.

\noindent
Note that whilst the above examples have the fixed distribution coming from the previously-learned policy of the first agent, this is not necessary as the algorithms merely require a distribution as an input (without any qualification on how that distribution was obtained). The usefuleness of the algorithms are, however, particularly clear when the (fixed) input distribution comes from a learned policy as in that case, the behaviours could be quite complex, and knowing \emph{a priori} how to influence the learning (e.g., through reward-shaping), can be difficult, if not impossible.

\section{Entropy-regularised Wasserstein Distance}

\noindent
Let $\mu$ and $\nu$ be two distributions with support $x_1, \cdots, x_{k_1}$ and $y_1, \cdots, y_{k_2}$ with $x_i$ and $y_j$ elements of a metric space $M$ for all $i,j$. The Wasserstein distance $W(\mu,\nu )$ between $P$ and $Q$ is defined as:
\begin{equation}
    W(\mu,\nu) := \min_{\kappa \in \mathbf{K}(\mu, \nu)}\langle \kappa , C \rangle
\end{equation}
\noindent

Where $C \in \mathbb{R}^{k_1} \times \mathbb{R}^{k_2}$ and satisfies $C_{i,j} = d(x_i, y_j)$ and $\mathbb{K}(\mu, \nu)$ denotes the set of couplings - joint distributions having $\mu$ and $\nu$ as left and right marginals respectively. Computing the Wasserstein distance and optimal coupling between two distributions $\mu$ and $\nu$  can be expensive. Instead, as proposed in ~\citet{cuturi2013sinkhorn}, a computationally friendlier alternative can be found in the entropy regularised variation of the Wasserstein distance, which for discrete distributions takes the form:
\begin{equation}\label{ent_reg_wass_disc}
W_\rho(\mu, \nu):= \min_{\kappa \in \mathbf{K}(\mu, \nu)} \quad \langle \kappa, C \rangle - \rho H(\kappa)
\end{equation}
 Here $H(\kappa) = -\sum_{i, j}\kappa_{ij} \log \kappa_{ij}$ is the entropy of the coupling $\kappa$, and $\rho>0$ is a regularisation parameter. 

\noindent 
More generally for the case of continuous support distributions, and following~\citet{genevay2016stochastic}, let $\mathcal{X}$ and $\mathcal{Y}$ be two metric spaces. Let $\mathcal{C}(\mathcal{X})$ be the space of real-valued continuous functions on $\mathcal{X}$ and let $\mathcal{M}_+^1(\mathcal{X})$ be the set of positive Radon measures on $\mathcal{X}$. Let $\mu \in \mathcal{M}_+^1(\mathcal{X})$ and $\nu \in \mathcal{M}_+^1(\mathcal{Y})$. Let $\mathbf{K}(\mu, \nu)$ be the set of couplings between $\mu, \nu$:
\begin{align*}
\mathbf{K}(\mu, \nu) &:= \{\kappa \in \mathcal{M}^1_+(\mathcal{X}\times \mathcal{Y})\, ; \, \forall (A, B) \subset \mathcal{X}\times \mathcal{Y}, \kappa (A \times \mathcal{Y}) = \mu(A), \kappa(\mathcal{X} \times B) = \nu(B)\}
\end{align*}

\noindent
That is the set of joint distributions $\kappa \in \mathcal{M}_+^1(\mathcal{X}\times \mathcal{Y})$ whose marginals over $\mathcal{X}$ and $\mathcal{Y}$ agree with $\mu$ and $\nu$ respectively.
Given a cost function $c \in \mathcal{C}(\mathcal{X} \times \mathcal{Y})$, the entropy-regularised Wasserstein distance $W_\rho(\mu, \nu)$ between $\mu$ and $\nu$ is defined as:
\begin{equation}\label{eq::smoothed_wass}
W_\rho(\mu, \nu):= \min_{\kappa \in \mathbf{K}(\mu, \nu)} \int_{\mathcal{X} \times \mathcal{Y}} c(x,y)d\kappa(x,y) + \rho \mathrm{KL}(\kappa || \mu \otimes \nu)
\end{equation}
where  $\forall (\kappa,  \xi) \in \mathcal{M}^1_+(\mathcal{X} \times \mathcal{Y})^2$, the KL-divergence between $\kappa$ and $\xi$ is defined by
\[
\mathrm{KL}(\kappa || \xi) = \int_{\mathcal{X} \times \mathcal{Y}}\left(\log \left(\frac{\mathrm{d}\kappa}{\mathrm{d}\xi}(x,y)\right)-1\right)\mathrm{d}\kappa(x,y).
\]
Here $\frac{\mathrm{d}\kappa}{\mathrm{d}\xi}(x,y)$ is the relative density of $\kappa$ with respect to $\xi$, and we define $\mathrm{KL}(\kappa || \xi) = +\infty$ if $\kappa$ doesn't have a density with respect to $\xi$.


\noindent
Note, we say \emph{the} optimal coupling because the above is a strongly-convex problem, unlike the unregularised version. The algorithm in~\citet{cuturi2013sinkhorn} is based on finding the dual variables of the Lagrangian by applying Sinkhorn's matrix scaling algorithm~\citep{sinkhorn1967diagonal}, which is an iterative procedure with linear convergence. 

\noindent
Stochastic optimisation algorithms were presented in~\citet{genevay2016stochastic} for the cases where (i) $\mu, \nu$ are both discrete, (ii) when one is discrete and the other continuous, and (iii) where both are continuous. We give algorithms for all cases, but due to their similarity and lack of space, we defer all but the continuous-continuous case to the Appendix.

\section{Algorithm for Continuous-Continuous Measures}
\noindent
Recall $M$ is a metric space and $f: \Gamma \rightarrow M$. Let $\nu$ be a fixed measure over $M$. We parameterise our policy with a vector $\theta \in \Theta$ where $\Theta$ is a parameter space. The objective is:
\begin{equation}
\max_{\theta \in \Theta} \, \,  V(\theta) + \lambda W_\rho(\mu_{\theta}, \nu) \label{eq::disc_objective}
\end{equation}
 where $V(\theta) := \mathbb{E}_{\pi_\theta}\left[\sum_{t \geq 0}\gamma^{t}r(s_t, a_t) \, |\, s_0\right] \equiv \mathbb{E}_{\tau \sim \pi_\theta}\left[R(\tau) \right]$ is the standard objective in RL,  $\mu_{\theta}$ is the distribution over $M$ induced by $\pi_\theta$ and $\lambda \in \mathbb{R}$ is a regularisation parameter. \textbf{\emph{{Note: $\lambda$ can be positive or negative. If it is positive then repulsion is promoted, whilst if it is negative, then  attraction is promoted.}}}

\noindent
In~\citet{genevay2016stochastic}, a stochastic optimisation
algorithm is presented for computing Wasserstein distance between continuous distributions. The dual formulation gives rise to test functions $(u,v) \in \mathcal{H}\times \mathcal{H}$ where $\mathcal{H}$ is a reproducing kernel Hilbert space (RKHS). The type of RKHS we will use will be generated by universal kernels~\citep{micchelli2006universal}, thereby allowing uniform approximability to continuous functions $u, v$. 
\begin{proposition}[Dual formulation~\citep{genevay2016stochastic}]\label{eq::smoothed_dual_wass}
\begin{align*}W_\rho( \mu, \nu) =\max_{u \in \mathcal{C}(\mathcal{X}), v \in \mathcal{C}(\mathcal{Y})}   \int_{\mathcal{X}} u(x) d\mu(x) + \int_{\mathcal{Y}} v(y) d\nu(y) \, - \rho \int_{\mathcal{X} \times \mathcal{Y}} \exp\left\{ \frac{u(x) + v(y) - c(x,y)}{\rho}    \right\}d\mu(x) d\nu(y)
\end{align*}
The solution $\kappa$ of problem \eqref{eq::smoothed_wass} can be recovered from a solution to the above by setting $d\kappa(x,y) = \exp\left\{ \frac{u(x) + v(y) - c(x,y)}{\rho}      \right\}d\mu(x) d\nu(y)$.
\end{proposition}

\noindent
Applying Proposition \ref{eq::smoothed_dual_wass} with $\mathcal{X} = \mathcal{Y} =	M$, we can write \eqref{eq::disc_objective} as
\begin{equation*}
\max_{u,v \in \mathcal{C}(M)} \, \max_{\theta \in \Theta} \, \mathbb{E}_{\tau \sim \pi_\theta, y \sim \nu} \left[\lambda \cdot F_\rho(f(\tau), y, u, v)  + R(\tau) \right]
\end{equation*}
where $F_\rho(x, y, u, v) := u(x) + v(y) -\rho \exp\left\{\frac{u(x) + v(y)-c(x,y)}{\rho}\right\}$.

\noindent
We consider functions $u,v$ to be elements of the RKHS $\mathcal{H}$ generated by $\mathcal{K}$. If $\mathcal{K}$ is a universal kernel over $\mathcal{X}$ ($\equiv M$), the search space for $u,v$ space will be rich enough to capture $\mathcal{C}(\mathcal{X})$~\citep{steinwart2008support}. Under these assumptions the $k$'th step of stochastic gradient ascent operation in the RKHS for $u,v$ takes the form:
\begin{equation}
(u_k,v_k) =  (u_{k-1}, v_{k-1}) + \frac{\text{constant}}{\sqrt{k}} \nabla_{u,v} F_\rho(x, y, u, v) \label{iterates}
\end{equation}
where $(x,y)$ are sampled from the product measure of the two measures being compared, in our case, $\mu_{\theta_k}$ and $\nu$. The implementation via kernels is through the following result:
\begin{proposition}[\cite{genevay2016stochastic}]\label{RKHS_prop}
The iterates $(u_k, v_k)$  defined in \eqref{iterates} satisfy $(u_k, v_k) = \sum_{i=1}^k\alpha_i(\kappa(\cdot, x_i), \kappa(\cdot, y_i))$, 
where $\alpha_i := \Pi_{B_r}\left(\frac{\text{const}}{\sqrt{i}} \left(1-\exp\left\{\frac{u_{i-1}(x_i) + v_{i-1}(y_i) - c(x_i, y_i)}{\rho}\right\}\right)\right)$, $(x_i, y_i)_{i=1}^k$ are i.i.d. samples from $\mu \otimes \nu$, and $\Pi_{B_r}$ is the projection on the centered ball of radius $r$. If the solutions of \eqref{eq::smoothed_dual_wass} are in $\mathcal{H} \times \mathcal{H}$ and if $r$ is large enough, then 
the iterates $(u_k, v_k)$ converge to a solution of \eqref{eq::smoothed_dual_wass}. 
\end{proposition}
To get Algorithm \ref{Alg:finite_cont_dual}, we note that by standard arguments~\citep[see e.g.,][]{sutton1998reinforcement},
\begin{align}
&\nabla_\theta\mathbb{E}_{\tau \sim \pi_\theta, y \sim \nu} \left[\lambda \cdot F_\rho(f(\tau), y, u, v)  + R(\tau) \right] \\
&= \mathbb{E}_{\tau \sim \pi_\theta, y \sim \nu} \left[\left(\lambda u(f(\tau))-\lambda\rho \exp\left\{\frac{u(f(\tau)) + v(y)-c(f(\tau),y)}{\rho}\right\}+ R(\tau)\right) \cdot \sum_{t \geq 0} \nabla_\theta \log \pi_{\theta} (a_t^{(\tau)} \, | \, s_t^{(\tau)})\right]. \label{Exxx}
\end{align}

Algorithm \ref{Alg:finite_cont_dual} presented below exploits Proposition \ref{RKHS_prop} to perform stochastic gradient decent on the policy parameter $\theta_i$ by sampling as a substitute for the expectation in \eqref{Exxx}. Parameters $(\alpha^{\theta}_i)_i$ define the learning rate. With each iteration, the algorithm is growing its estimate of the functions $u$ and $v$ in the RKHS and evaluating them in lines \eqref{ui} and \eqref{vi} using the previous samples $X_j, Y_j$ to define the basis functions $\kappa(\cdot, X_j), \kappa(\cdot, Y_j)$. The variable $Z$ is merely for notational convenience. 

\begin{algorithm}[H]
\textbf{Input: $\theta_0, f, M, \lambda, \rho, \nu, (\alpha^{\theta}_i)_i$}\;
\textbf{Initialise: $u_0, v_0, \theta_0$}\;
 \For{$i=1, 2, \ldots$}{
  sample $\tau \sim \pi_{\theta_{i-1}}$\;

	$X_i \leftarrow f(\tau)$\;

	sample $Y_i \sim \nu$\;

	$u_{i-1}(X_i) := \sum_{j=1}^{i-1}\alpha_j\kappa(X_i, X_j)$\; \label{ui}

	$v_{i-1}(Y_i) := \sum_{j=1}^{i-1}\alpha_j\kappa(Y_i, Y_j)$\; \label{vi}
	
	$Z \leftarrow \exp\left\{\frac{u_{i-1}(X_i) + v_{i-1}(Y_i)-c(X_i,Y_i)}{\rho}\right\}$\;
	
	$\theta_{i} \leftarrow \theta_{i-1} + \alpha^{\theta}_i \cdot \left[ \left(\lambda u_{i-1}(X_i)-\lambda\rho Z + R(\tau)\right) \cdot \sum_{t \geq 0} \nabla_\theta \log \pi_{\theta_{i-1}} (a_t^{(\tau)} \, | \, s_t^{(\tau)})\right]$\;

	$\alpha_i := \frac{\text{constant}}{\sqrt{i}} \left(1-Z\right)$
 }
 \caption{Stochastic gradient for continuous measures}
\label{Alg:finite_cont_dual}
\end{algorithm}

\section{Experiments}
\subsection{Testing for an Attractive Scenario}
\begin{wrapfigure}{r}{0.25\textwidth}
  \vspace{-20pt}
  \begin{center}
    \includegraphics[width=0.25\textwidth]{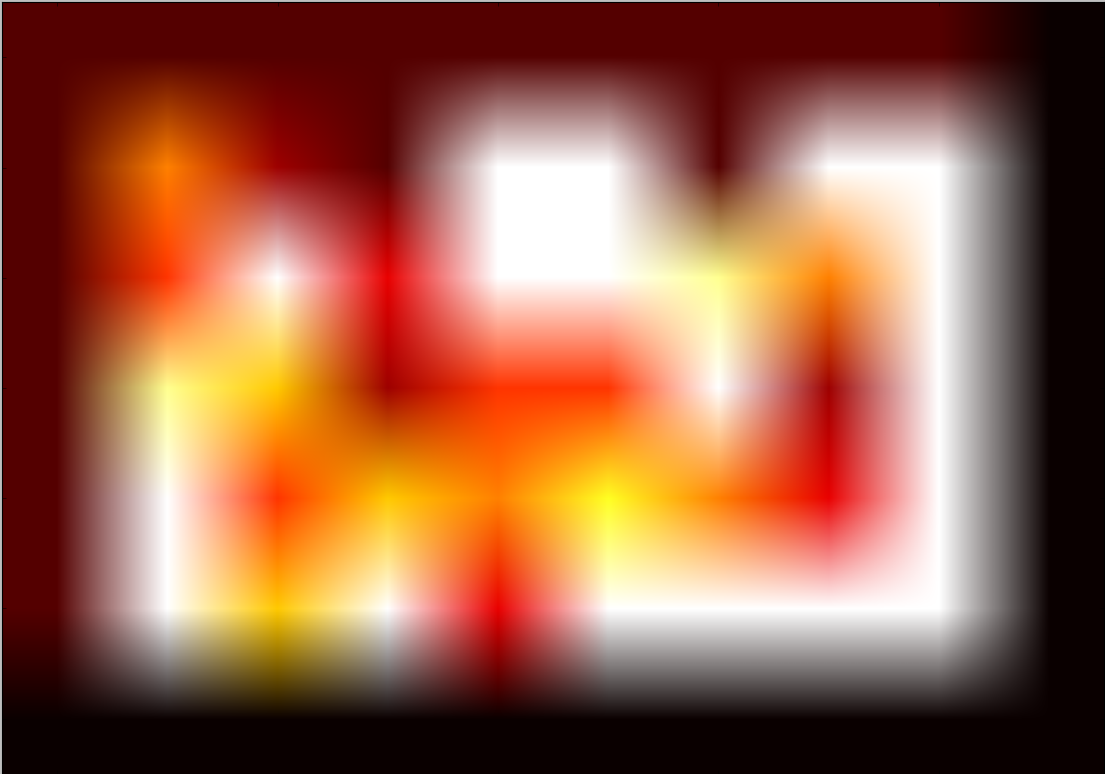}
  \end{center}
	  \vspace{-20pt}
  \caption{Gridworld. Darker is cheaper.}\label{grid_height}
  \vspace{-20pt}
\end{wrapfigure}

\noindent
We have a 7x10 gridworld with a non-negative integer ``height'' associated to each cell, Figure \ref{grid_height}. An agent starts in the lower-left corner cell and there is an absorbing state on the upper-right corner cell. Each movement incurs a penalty of $-1-z$ where $z$ is the height of the cell moved to. An episode is terminated either by a time-out or reaching the absorbing state. 
\noindent
The trajectory mapping $f(\tau)$ is a probability distribution of cell visits made by $\tau$, i.e., a count is made of the number of times each cell is visited and this value is normalised by the trajectory length. Thus, $f(\tau)$ is a point in the $7 \times 10 -1= 69$ dimensional probability simplex. We set $\nu$ to be Dirac measure on the unique optimal solution. Policy parameterisation is by radial basis functions centred on each cell.

\begin{figure}
\centering
\begin{subfigure}[b]{0.32\textwidth}
    \includegraphics[width=\textwidth]{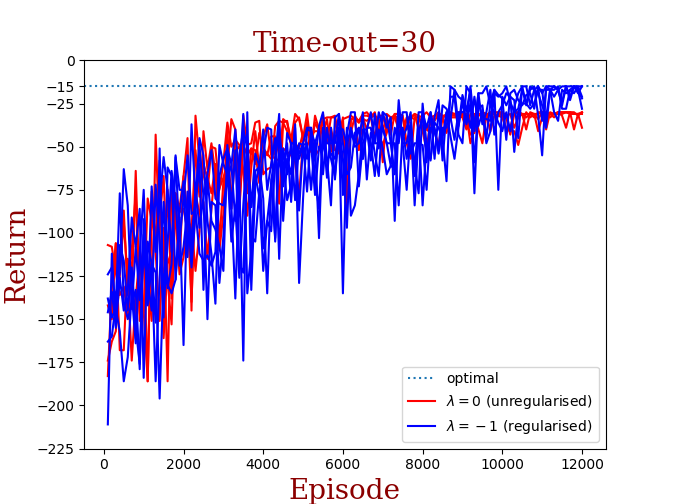}
\end{subfigure}
\begin{subfigure}[b]{0.32\textwidth}
		\includegraphics[width=\textwidth]{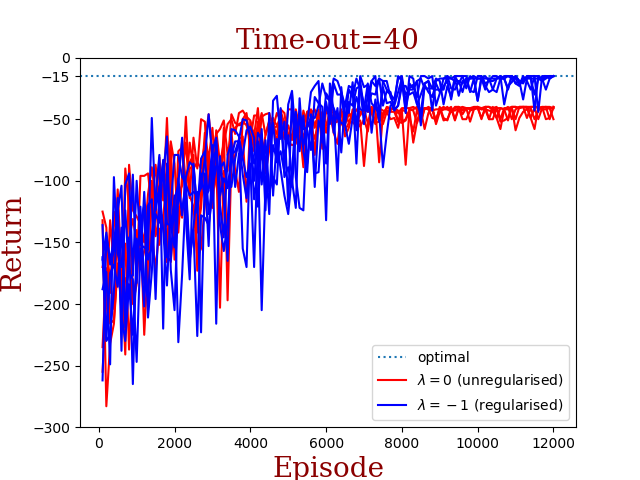}
\end{subfigure}
\begin{subfigure}[b]{0.32\textwidth}
		\includegraphics[width=\textwidth]{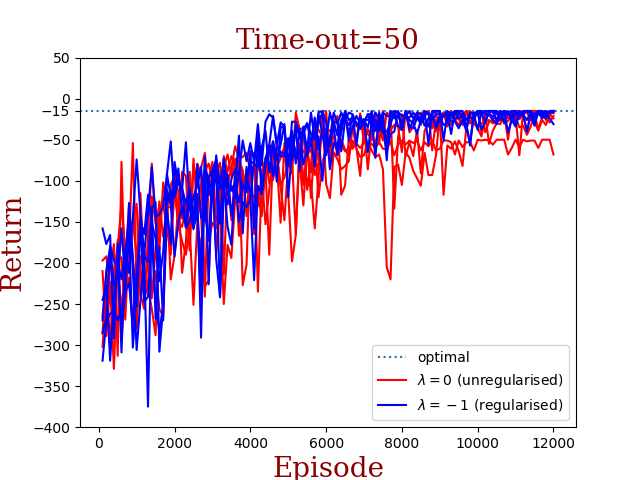}
\end{subfigure}
    \caption{Attractive scenario on gridworld; regularised vs unregularised. Time-out at 30, 40 and 50 steps. Unregularised finds optimal solution more slowly or never finds it.}\label{SDresults}
\end{figure}

\noindent
For this set of experiments, the aim was to test the effectiveness of our algorithms with $\lambda=-1$ against policy gradient without Wasserstein regularisation (by setting $\lambda=0$). Specifically, we wanted to determine if our algorithms got better returns for a given number of episodes (i.e., iterations). In all cases, the entropy regulariser $\rho=1$. We performed three sets of experiments based on the time-out, i.e., maximum length of a trajectory before termination of an episode: 30 steps, 40 steps and 50 steps. If the agent did not reach the absorbing
state before the time-out, it would get a penalty. Hence, an optimal trajectory would incur a total cost of $-15$. 
Each experiment consisted of 12,000
episodes and the result recorded was the return on every 100'th episode. For each time out, we ran five experiments for each of $\lambda=-1$ and $\lambda=0$. As can be seen from Figure \ref{SDresults}, $\lambda=-1$ (blue) out-performed $\lambda=0$ (red) in general. Indeed, the former often found a (nearly) optimal policy whereas the latter often did not. Clock execution
time was not hindered by $\lambda=-1$, indeed, because good policies were found more quickly, it was usually quicker to complete the experiments.

\subsection{Testing for a Repulsive Scenario}
\noindent
The environment is set on a two dimensional plane where there are two goals (marked with a dot). The state space equals the $(x,y)$ positions of the agent and the reward varies in an inversely proportional way to the distance between the agent and the closest goal. The desired objective is to find the two qualitatively distinct optimal policies in an automated way. The algorithm that starts with two randomly initialized neural networks each with two hidden layers of 15 nodes each. The metric mapping is the $x$ position of the agent along the given trajectory. In contrast to the algorithms described in the previous sections, we do not start with a target distribution $\nu$ to repel at the start of the procedure. Instead, the algorithm is able to dynamically guide exploration and find two distinct policies. 

\begin{figure}[h]
  \begin{minipage}[b]{0.18\linewidth}
    \centering
    \includegraphics[width=\linewidth]{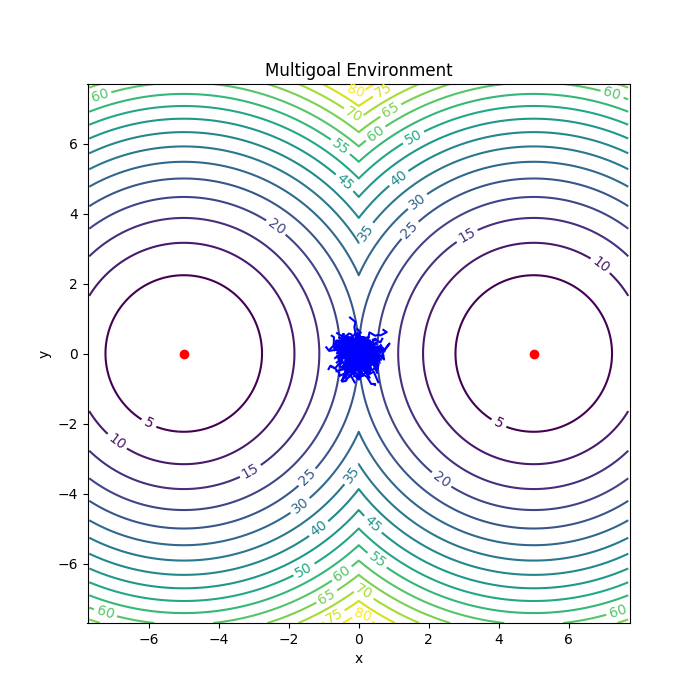} 
  \end{minipage}
  \begin{minipage}[b]{0.18\linewidth}
    \centering
    \includegraphics[width=\linewidth]{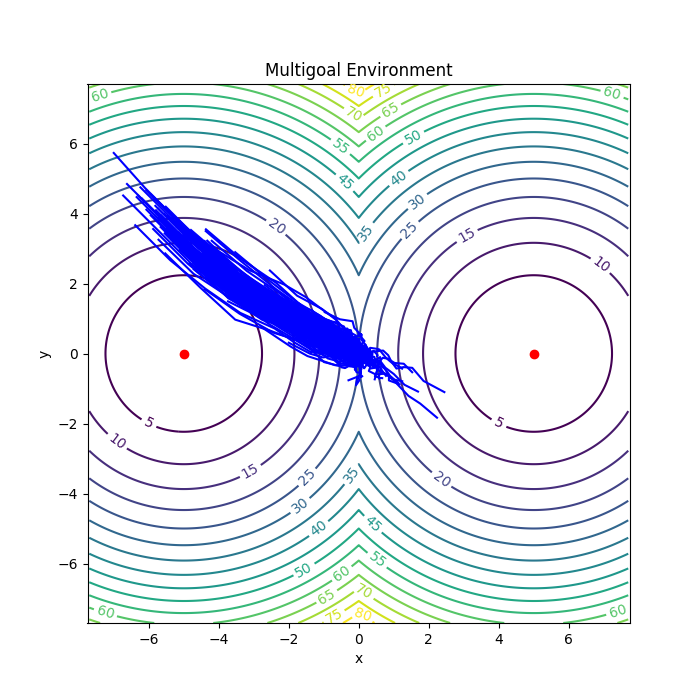} 
  \end{minipage} 
	 \begin{minipage}[b]{0.18\linewidth}
    \centering
    \includegraphics[width=\linewidth]{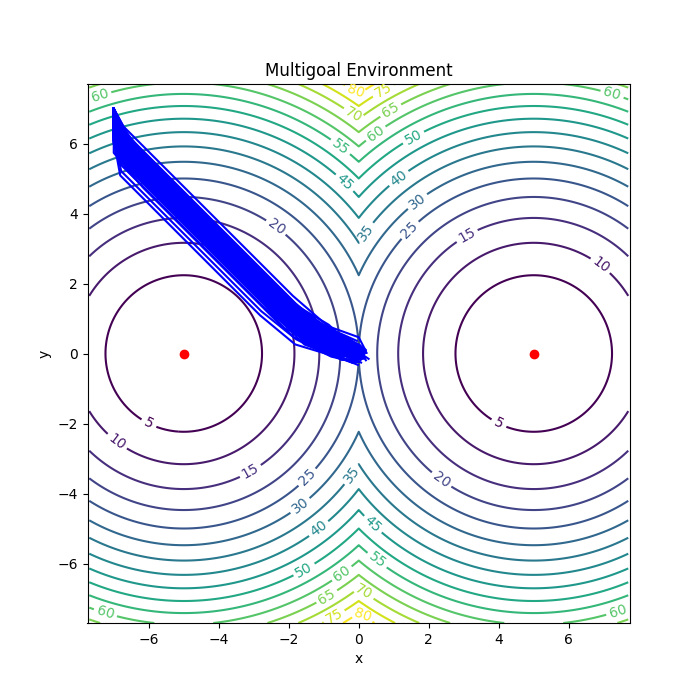} 
  \end{minipage} 
		 \begin{minipage}[b]{0.18\linewidth}
    \centering
    \includegraphics[width=\linewidth]{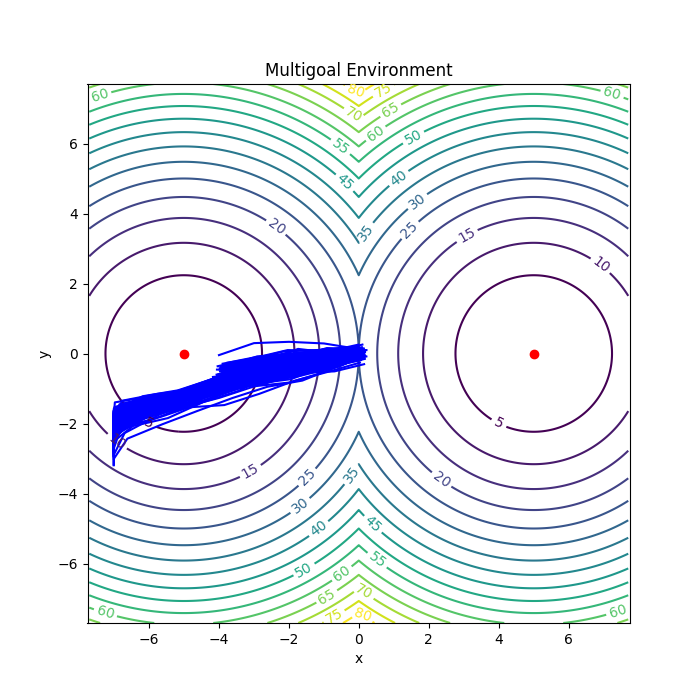} 
  \end{minipage} 
		 \begin{minipage}[b]{0.18\linewidth}
    \centering
    \includegraphics[width=\linewidth]{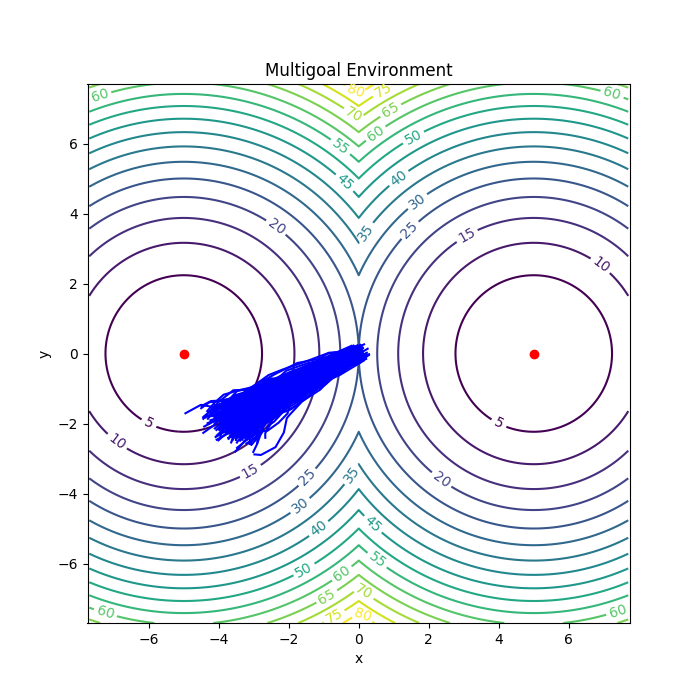} 
  \end{minipage} 
	
	\begin{minipage}[b]{0.18\linewidth}
    \centering
    \includegraphics[width=\linewidth]{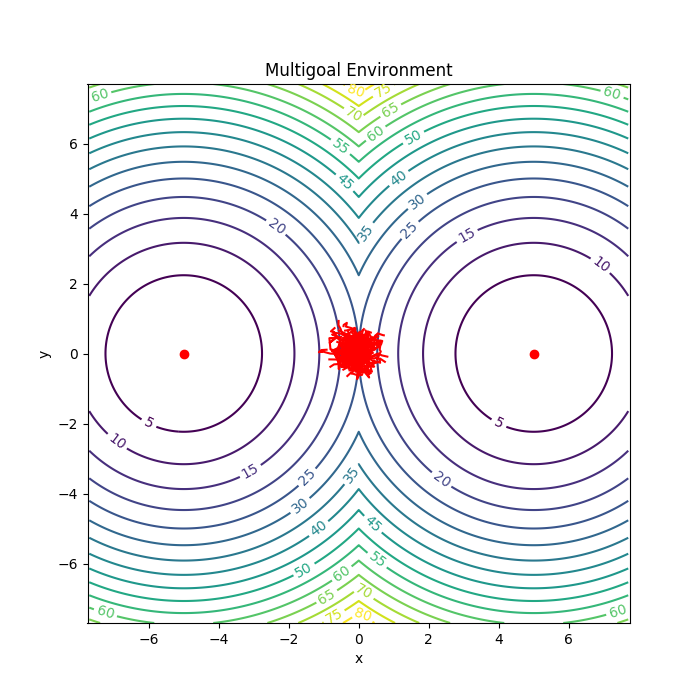} 
  \end{minipage}
  \begin{minipage}[b]{0.18\linewidth}
    \centering
    \includegraphics[width=\linewidth]{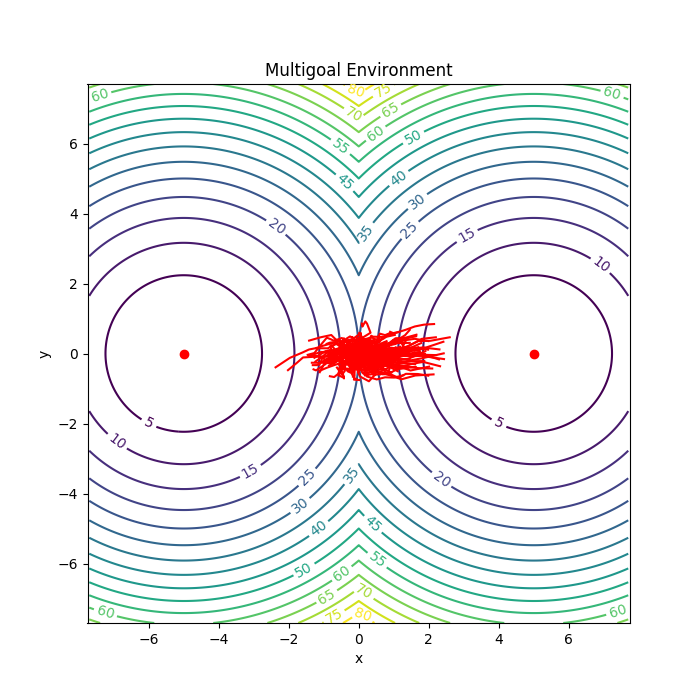} 
  \end{minipage} 
	 \begin{minipage}[b]{0.18\linewidth}
    \centering
    \includegraphics[width=\linewidth]{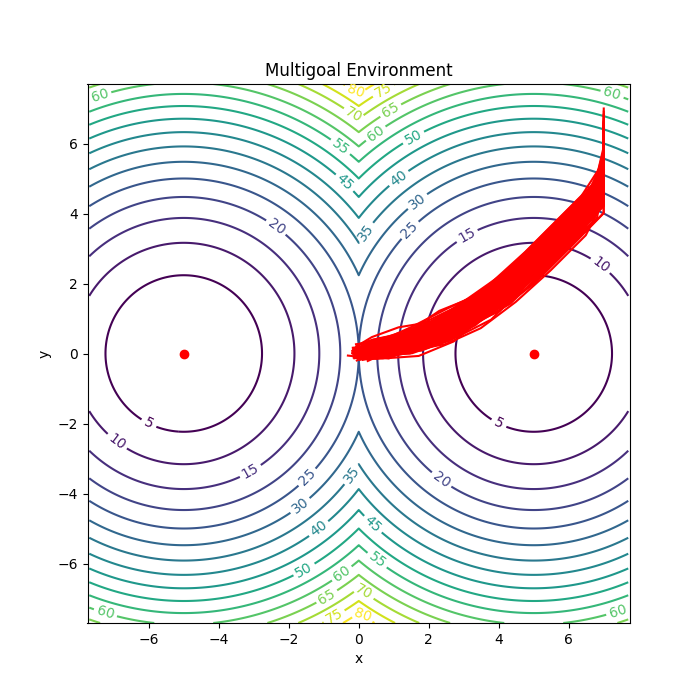} 
  \end{minipage} 
		 \begin{minipage}[b]{0.18\linewidth}
    \centering
    \includegraphics[width=\linewidth]{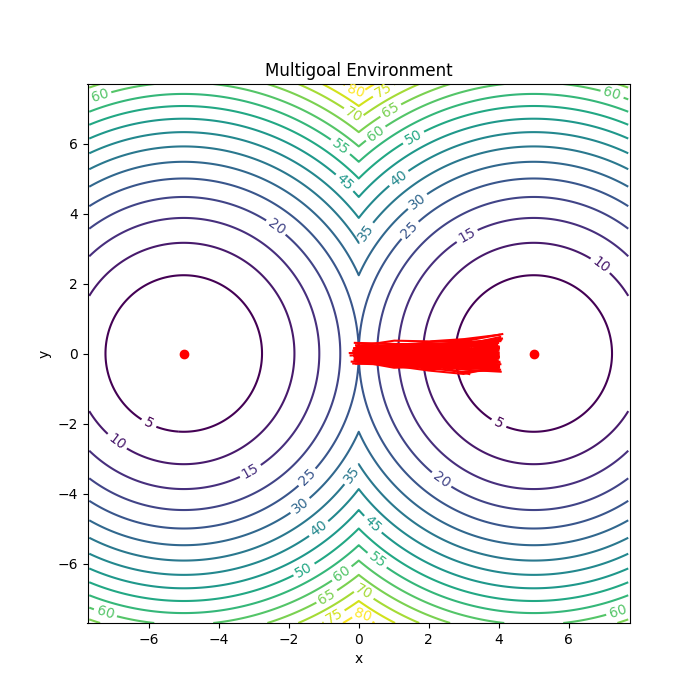} 
  \end{minipage} 
		 \begin{minipage}[b]{0.18\linewidth}
    \centering
    \includegraphics[width=\linewidth]{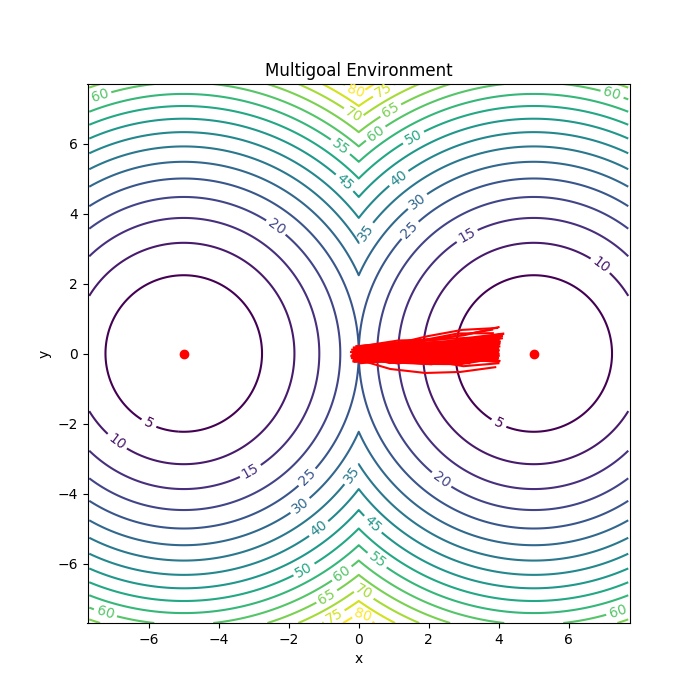} 
  \end{minipage}

		\begin{minipage}[b]{0.18\linewidth}
    \centering
    \includegraphics[width=\linewidth]{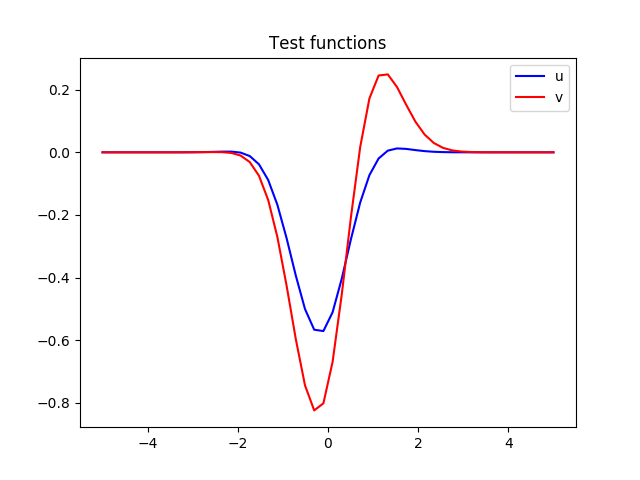} 
  \end{minipage}
  \begin{minipage}[b]{0.18\linewidth}
    \centering
    \includegraphics[width=\linewidth]{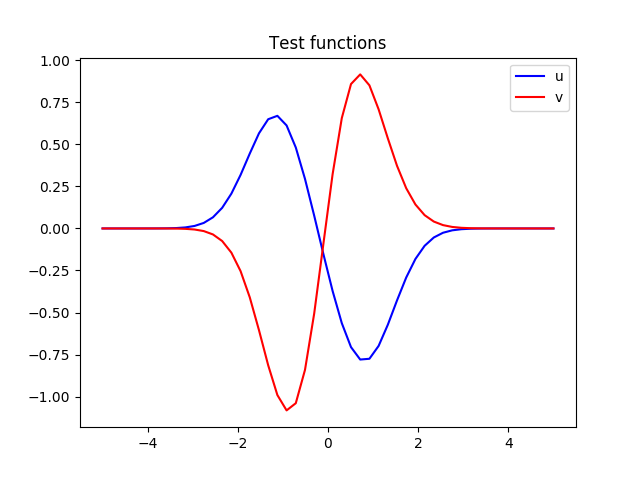} 
  \end{minipage} 
	 \begin{minipage}[b]{0.18\linewidth}
    \centering
    \includegraphics[width=\linewidth]{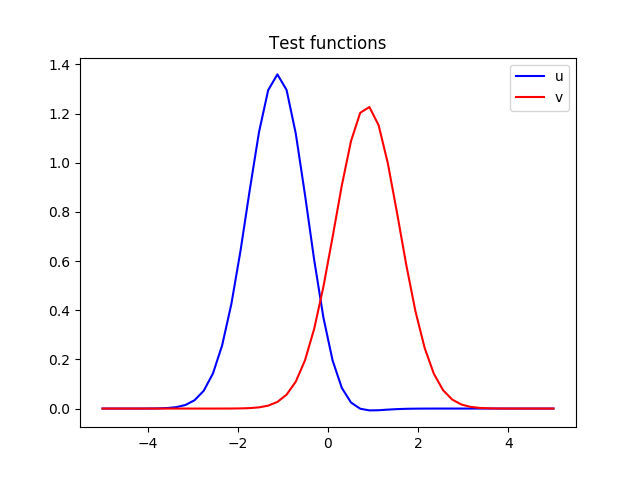} 
  \end{minipage} 
		 \begin{minipage}[b]{0.18\linewidth}
    \centering
    \includegraphics[width=\linewidth]{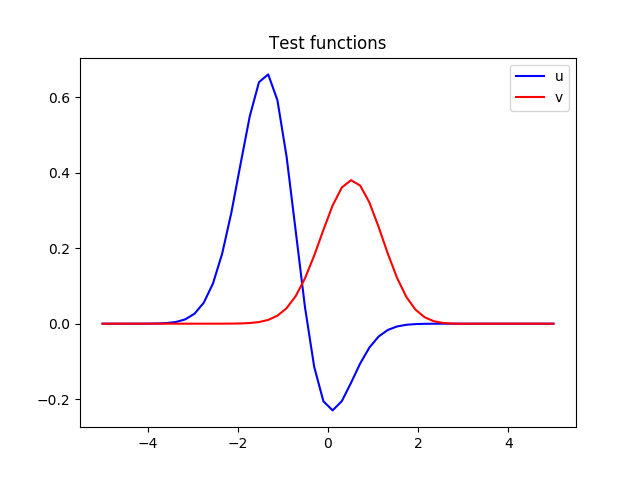} 
  \end{minipage} 
		 \begin{minipage}[b]{0.18\linewidth}
    \centering
    \includegraphics[width=\linewidth]{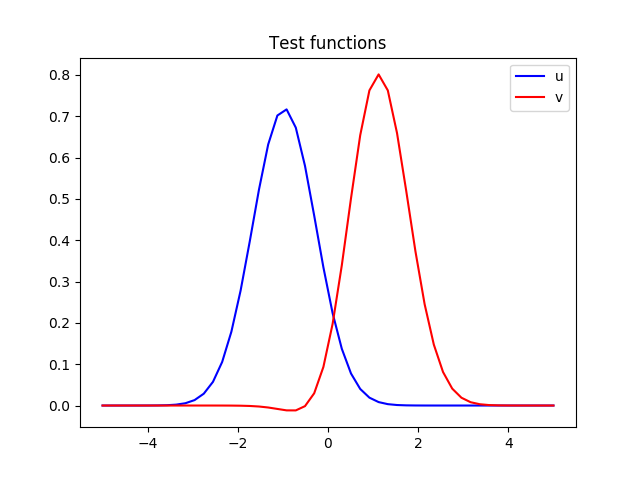} 
  \end{minipage} 
	
	\begin{minipage}[b]{0.18\linewidth}
    \centering
    Iteration 0
  \end{minipage}
  \begin{minipage}[b]{0.18\linewidth}
    \centering
    Iteration 15
  \end{minipage} 
	 \begin{minipage}[b]{0.18\linewidth}
    \centering
    Iteration 30
  \end{minipage} 
	\begin{minipage}[b]{0.18\linewidth}
    \centering
    Iteration 60
  \end{minipage} 
		 \begin{minipage}[b]{0.18\linewidth}
    \centering
    Iteration 100
  \end{minipage}
	\caption{Learning two policies by repulsive Wasserstein regularisation.}\label{fig7} 
\end{figure}

\noindent
The results of our runs are shown in Figure \ref{fig7}. Each iteration represents a policy gradient step in the parameter space of each policy. In order to compute these gradient estimates, and to find the test functions $u,v$, we use 100 rollouts of each policy in each iteration. We use $\rho=0.01$. Good convergence to two distinct policies is achieved after roughly 100 iterations. In the following plots we show, along with sample trajectories from each of the agents at a particular iteration number, images of the test functions $u,v$ and their evolution through time. It can be seen how these modify the reward structure as the algorithm runs to penalise/reward trajectories in opposing ways between the two agents. The images corresponding to iteration 15 are particularly telling as they show how even before the agents commit to a specific direction the test scores produced by our procedure strongly favour diversity. 

\vspace{-5mm}

\section{Conclusion, Related Work and Future Work}
\noindent
We have introduced the notion of Wasserstein distance regularisation into reinforcement learning as a means to quantitatively compare the trajectories (i.e., \emph{behaviours}) of different policies. To the best of our knowledge, this paper is the first such example of this application. Depending on the sign of
the regulariser, this technique allows policies to diverge  or converge in their behaviour. This has been demonstrated through testing of algorithms presented in this paper. For future work, it would, perhaps, be natural to compare our techniques to those of imitation and inverse reinforcement learning~\citep{abbeel2004apprenticeship, argall2009survey}, or techniques like guided policy search~\citep{levine2013guided}. Direct comparison is not immediate, since our technique obliges the user to define the metric mapping, but we believe this extra demand would
pay off in learning rates or through other considerations such as the flexibility it gives the user to decide what the important features of behaviour are, and tailor learning to them.

\newpage
\bibliography{ref}

\appendix
\section*{Appendix A. Wasserstein-regularised RL for (Semi-)Discrete Measures}
\noindent
We discuss Wasserstein distance between probability measures $\mu, \nu$. Suppose $M$ is a metric space, and $\mu = \sum_{i=1}^n \mu_i\delta_{x_i}$, $\nu = \sum_{j=1}^m \nu_j\delta_{y_j}$ are finite discrete measures where $x_i, y_j \in M$. A coupling $\kappa=\kappa(\mu, \nu)$ of $\mu$ and $\nu$ is a measure over $\{x_1, \ldots, x_n\} \times \{y_1, \ldots y_m\}$ that preserves marginals, i.e, $\mu_i=\sum_j\kappa(\mu_i, \nu_j)$ $\forall i$ and $\nu_j = \sum_{i}\kappa(\mu_i, \nu_j)$ $\forall j$. This then induces a cost of ``moving'' the mass of $\mu$ to $\nu$, given as the (Frobenius) inner product $\langle \kappa, C \rangle$ where the matrix $C \in \mathbb{R}^{n \times m}$ has $[C]_{ij}=c_{ij}=d(x_i, y_j)$, i.e., the cost of moving a unit of measure from $x_i$ to $y_j$. Minimised over the space of all couplings $\mathbf{K}(\mu, \nu)$, we get the Wasserstein distance, also known as the \emph{Earth-Mover Distance} (EMD)~\citep{villani2008optimal, santambrogio2015optimal}. 

\noindent
Let $\mathbb{P}_n$ be the $n-1$ dimensional probability simplex. We also have a fixed distribution $\nu = (\nu_1, \ldots, \nu_n) \in \mathbb{P}_n$ over points $(y_1, \ldots, y_n) \in M^n$ i.e, with mild abuse of notation, the measure is $\mu=\sum_{j=1}^n\nu_j\delta_{y_j}$. Lastly, we have a cost matrix $C \in \mathbb{R}_+^{n \times n}$ Note $f, M, \nu, C$ are inputs to the algorithm.

\subsection*{Gradient-based Optimisation}
\noindent
Following \eqref{eq::disc_objective} we have:
\begin{equation*}
\nabla_\theta \left\{ V(\theta) + \lambda W_\rho(\mu_{\theta}, \nu)\right\} = \nabla_\theta V(\theta)+ \lambda \nabla_\mu W_\rho(\mu, \nu)\vert_{\mu=\mu_\theta} \cdot \nabla_\theta\mu_\theta   
\end{equation*}
Per the standard policy gradient approach~\citep{sutton1998reinforcement}, we can sample trajectories to get an unbiased estimate of $\nabla_\theta V(\theta)$. Indeed, for any
function $g: \Gamma \rightarrow \mathbb{R}$, 
\begin{equation}
\nabla_\theta \mathbb{E}_{\tau \sim \pi_\theta}\left [ g(\tau) \right] = \mathbb{E}_{\tau \sim \pi_\theta}\left [ g(\tau)\, \sum_{t \geq 0}\nabla_\theta \log \pi_\theta (a_t \, |, \, s_t) \right], 
\label{eq:grad_swap}
\end{equation}
meaning we can sample trajectories to obtain unbiased estimates of $\nabla_\theta\mu_\theta$. 

\noindent
Finally, the term $\nabla_\mu W_\rho(\mu, \nu)\vert_{\mu=\mu_\theta}$ can be dealt with using the Sinkhorn algorithm itself: In the computation of $W_\rho(\mu, \nu)$ for a given pair $\mu, \nu$, the algorithm computes their optimal dual variables $u^*, v^*$, respectively. It can do so because, as mentioned above, the entropy-regularisation makes the optimisation strongly convex and strong duality is exhibited. Then $u^*$ is a (sub)gradient of  $W_\rho(\mu, \nu)$ with respect to $\mu$ (this will be discussed further below). Thus, given an estimate of $\mu_\theta$, we can estimate $\nabla_\mu W_\rho(\mu, \nu)\vert_{\mu=\mu_\theta}$. 

\noindent
Putting it all together, we have derived a simple stochastic gradient algorithm. In Algorithm \ref{Alg:finite_disc}, $(\alpha_i)_i$ is a learning rate.

\begin{algorithm}[H]
 \textbf{Input: $\theta_0, f, M, \lambda, \rho, \nu, (\alpha_i)_i$}\;
\textbf{Initialise: $\theta_0$}\;
 \For{$i=1, 2, \ldots$}{
  sample $\tau \sim \pi_{\theta_{i-1}}$\;
	compute estimate $gtv(\tau)$ of $\nabla_\theta V(\theta)\vert_{\theta=\theta_{i-1}}$  using \eqref{eq:grad_swap}\; 
	compute estimate $gtm(\tau)$ of $\nabla_\theta \mu_\theta\vert_{\theta=\theta_{i-1}}$ using \eqref{eq:grad_swap}\;
	update estimate $\hat{\mu_\theta}$ of $\mu_\theta$ using $\tau$\;
	compute estimate $gtw$ of $\nabla_\mu W_\rho(\mu, \nu)\vert_{\mu=\mu_{\theta_{i-1}}}$ using Sinkhorn ~\cite{cuturi2013sinkhorn} and $\hat{\mu_\theta}$\;
	$\theta_{i} \leftarrow \theta_{i-1} + \alpha_i \cdot \left(gtv(\tau) + \lambda \cdot gtw \cdot gtm(\tau)\right)$
 }
 \caption{Wasserstein RL for finite discrete measures}
\label{Alg:finite_disc}
\end{algorithm}

\subsection*{Stochastic Alternating Optimisation via Dual Formulation}
The dual of the primal problem \eqref{ent_reg_wass_disc} was studied in~\citet{cuturi2014fast}. Applying it, we get the following equivalent of \eqref{eq::disc_objective}:
\begin{equation}
\max_{\theta \in \Theta} \, \max_{u, v \in \mathbb{R}^n} \,  \lambda \left( \langle u, \mu_\theta \rangle + \langle v, \nu \rangle - \rho B(u,v) \right) + V(\theta) \label{eq:maxmax1}
\end{equation}
where
\begin{equation*}
B(u,v) := \sum_{i,j}\exp\left\{\frac{u_i+v_j-c_{ij}}{\rho}\right\}.
\end{equation*}

\noindent
Swapping the order of maximisations:
\begin{equation*}
 \max_{u, v \in \mathbb{R}^n} \, \lambda \left(\langle v, \nu \rangle - B(u,v) \right) + \max_{\theta \in \Theta} \, \lambda \langle u, \mu_\theta \rangle + V(\theta).
\end{equation*}

\noindent
The term $\langle u, \mu_\theta \rangle$ is an expectation, and the above can be re-written:
\begin{equation*}
 \max_{u, v \in \mathbb{R}^n} \, \lambda \left(\langle v, \nu \rangle - B(u,v) \right) + \max_{\theta \in \Theta}\mathbb{E}_{\tau \sim \pi_\theta}\left[ \lambda u(f(\tau))+ R(\tau)\right]
\end{equation*}
for an appropriate function $u$. 

\noindent
An iterative algorithm can proceed by alternatively fixing $u, v$ and maximising $\theta$, and vice versa. When $u$ is fixed, we can apply policy gradient~\citep{sutton1998reinforcement} to the term $\mathbb{E}_{\tau \sim \pi_\theta}\left[ \lambda u(f(\tau))+ R(\tau)\right]$; 
\begin{equation}
\nabla_\theta \mathbb{E}_{\tau \sim \pi_\theta}\left[ \lambda u(f(\tau))+ R(\tau)\right]\nonumber = \mathbb{E}_{\tau \sim \pi_\theta}\left[ \left(\lambda u(f(\tau))+ R(\tau)\right) \cdot \sum_{t \geq 0} \nabla_\theta \log \pi_\theta (a_t^{(\tau)} \, | \, s_t^{(\tau)})\right] 
\label{eq:Efunsumgradlogpi}
\end{equation}
\noindent
Thus, sampling a trajectory from $\pi_\theta$ and using it to compute the bracketed term in \eqref{eq:Efunsumgradlogpi} gives an unbiased estimate of the true gradient. This can be used to update $\theta$. Further, observe that fixing $\theta$, the expression to be maximised in \eqref{eq:maxmax1} is differentiable in $u,v$. This provides the means to increase $u, v$. This iterative alternating maximisation procedure is summarised in Algorithm \ref{Alg:finite_disc_dual}. 

\begin{algorithm}[H]
\textbf{Input: $\theta_0, f, M, \lambda, \rho, \nu, (\alpha^{\theta}_i)_i, (\alpha^{u}_i)_i, (\alpha^{v}_i)_i$}\;
\textbf{Initialise: $u_0, v_0, \theta_0$}\;
 \For{$i=1, 2, \ldots$}{
  sample $\tau \sim \pi_{\theta_{i-1}}$\;
	$\theta_{i} \leftarrow \theta_{i-1} + \alpha^{\theta}_{i-1} \cdot \left[ \left(\lambda u_{i-1}(f(\tau))+ R(\tau)\right) \cdot \sum_{t \geq 0} \nabla_\theta \log \pi_{\theta_{i-1}} (a_t^{(\tau)} \, | \, s_t^{(\tau)})\right]$\;
	$u_i \leftarrow u_{i-1} + \alpha_i^u \cdot \lambda \cdot \left(f(\tau) - \nabla_u B(u, v) \vert_{u=u_{i-1}, v=v_{i-1}}\right)$\;
	sample $Y \sim \nu$\;
	$v_i \leftarrow v_{i-1} + \alpha_i^v \cdot \lambda \cdot \left(Y - \nabla_v B(u, v) \vert_{u=u_{i-1}, v=v_{i-1}}\right)$
 }
 \caption{Stochastic gradient for finite discrete measures, dual formulation}
\label{Alg:finite_disc_dual}
\end{algorithm}
\noindent
The advantage of this algorithm is that the distributions $\mu_\theta$ and $\nu$ do not need to be represented explicitly, they only have to be sampled from. 

\noindent
The above can be generalised to the case $\mu$ is an arbitrary measure $\nu = \sum_{j=1}^m \nu_j\delta_{y_j}$ remains discrete. Starting with the form of the Wasserstein distance given in \eqref{eq::smoothed_wass} (which is slightly different to the version defined in \eqref{ent_reg_wass_disc}), and taking $\nu$ as discrete, it was shown in~\citet{genevay2016stochastic} that by writing first-order optimality conditions, one gets:
\begin{equation*}
W_\rho(\mu, \nu) = \max_{v \in \mathbb{R}^m} \mathbb{E}_{X \sim \mu}\left[h(X,v)\right]
\end{equation*}
where
\begin{equation*}
h(x,v) :=\langle v, \nu \rangle - \rho \log \left(\sum_{j=1}^m \exp\left\{\frac{v_j-c(x,y_j)}{\rho}\right\}\nu_j\right).
\end{equation*}
Thus, in our case, we would have $f(\tau) \sim \pi_{\theta}$ in place of $X \sim \mu$ above, and our objective would be
\begin{eqnarray*}
\max_{\theta}\, \, V(\theta) + \lambda \max_{v \in \mathbb{R}^m} \mathbb{E}_{f(\tau) \sim \pi_{\theta}}\left[h(f(\tau),v)\right]= \max_{\theta}\max_{v \in \mathbb{R}^m} \, \,  +  \mathbb{E}_{f(\tau) \sim \pi_{\theta}}\left[R(\tau)+\lambda h(f(\tau),v)\right].
\end{eqnarray*}
Because of the structure of $\nabla_u h(u,v)$, we cannot use a sampled vector $Y \sim \nu$ as we did with Algorithm \ref{Alg:finite_disc_dual}, we have to access each element in the vector $\nu$.
Hence, this algorithm is useful when $m$ is not too large. With that said, we can use incremental alternating gradient ascent, as summarised in Algorithm \ref{AlgSemiDisc}. 

\begin{algorithm}[H]
\textbf{Input: $\theta_0, f, M, \lambda, \rho, \nu, (\alpha^{\theta}_i)_i, (\alpha^{v}_i)_i$}\;
\textbf{Initialise: $v_0, \theta_0$}\;
 \For{$i=1, 2, \ldots$}{
  sample $\tau \sim \pi_{\theta_{i-1}}$\;
	$\theta_{i} \leftarrow \theta_{i-1} + \alpha^{\theta}_i \cdot \left[ \left(\lambda h_\rho(f(\tau), v_{i-1})+ R(\tau)\right) \cdot \sum_{t \geq 0} \nabla_\theta \log \pi_{\theta_{i-1}} (a_t^{(\tau)} \, | \, s_t^{(\tau)})\right]$\;
	$v_i \leftarrow v_{i-1} + \alpha_i^v \cdot \lambda \cdot \nabla_v h(x, v) \vert_{x=f(\tau), v=v_{i-1}}$
 }
 \caption{Stochastic gradient for discrete $\nu$, arbitrary $\mu_\theta$}
\label{AlgSemiDisc}
\end{algorithm}

\end{document}